\documentclass[runningheads]{llncs}
\usepackage[T1]{fontenc}
\usepackage{graphicx,verbatim}
\usepackage{booktabs}       % Required for professional table lines (\toprule, \midrule, etc.)
\usepackage{multirow}       % Required for the multi-row cells in the comparison tables
\usepackage{siunitx}        % Essential for numerical alignment and scientific notation
\usepackage{amsmath}        % Required for the NMSE and mathematical equations
\usepackage{amssymb}        % Provides additional mathematical symbols
\usepackage{url}            % Handles URL formatting in the bibliography
\usepackage{hyperref}
\usepackage{float}
\usepackage{subcaption}
\usepackage{placeins}

\begin{document}
\title{Anatomical Token Uncertainty for Transformer-Guided Active MRI Acquisition}
%\titlerunning{Abbreviated paper title}
% If the paper title is too long for the running head, you can set
% an abbreviated paper title here
%
% \begin{comment}  %% Removed for anonymized MICCAI submission
\author{Lev Ayzenberg\inst{1} \and
Shady Abu-Hussein\inst{2} \and
Raja Giryes\inst{1} \and
Hayit Greenspan\inst{1}}
\authorrunning{L. Ayzenberg et al.}
% First names are abbreviated in the running head.
% If there are more than two authors, 'et al.' is used.
%
\institute{
Tel Aviv University, Faculty of Engineering, Tel Aviv, Israel \and
University of Cambridge, Department of Engineering, Cambridge, UK
}

% \end{comment}

% \author{Anonymized Authors}  %% Added for anonymized MICCAI submission
% \authorrunning{Anonymized Author et al.}
% \institute{Anonymized Affiliations \\
%     \email{email@anonymized.com}}
  
\maketitle              % typeset the header of the contribution
\begin{abstract}
Full data acquisition in MRI is inherently slow, which limits clinical throughput and increases patient discomfort. Compressed Sensing MRI (CS-MRI) seeks to accelerate acquisition by reconstructing images from under-sampled k-space data, requiring both an optimal sampling trajectory and a high-fidelity reconstruction model. In this work, we propose a novel active sampling framework that leverages the inherent discrete structure of a pretrained medical image tokenizer and a latent transformer. By representing anatomy through a dictionary of quantized visual tokens, the model provides a well-defined probability distribution over the latent space. We utilize this distribution to derive a principled uncertainty measure via token entropy, which guides the active sampling process. We introduce two strategies to exploit this latent uncertainty: (1) Latent Entropy Selection (LES), projecting patch-wise token entropy into the $k$-space domain to identify informative sampling lines, and (2) Gradient-based Entropy Optimization (GEO), which identifies regions of maximum uncertainty reduction via the $k$-space gradient of a total latent entropy loss.
We evaluate our framework on the fastMRI singlecoil Knee and Brain datasets at $\times 8$ and $\times 16$ acceleration. Our results demonstrate that our active policies outperform state-of-the-art baselines in perceptual metrics, and feature-based distances.
Our code is available at \url{https://github.com/levayz/TRUST-MRI}.

\keywords{Active Sampling \and Compressed Sensing \and Transformer \and MRI Reconstruction}
% Authors must provide keywords and are not allowed to remove this Keyword section.

\end{abstract}
\section{Introduction}
Magnetic Resonance Imaging (MRI) acquires data in the spatial Fourier domain, namely the $k$-space, where each measurement corresponds to a specific spatial frequency component of the image. Full $k$-space acquisition is inherently slow, limiting clinical throughput and increasing patient discomfort.
Compressed Sensing (CS)~\cite{cs2,cs} enables accelerated MRI by reconstructing images from undersampled $k$-space measurements. 
Formally, let $\mathbf{x} \in \mathbb{C}^N$ denote the ground-truth image and $\mathbf{y} \in \mathbb{C}^N$ the acquired measurements. The acquisition procedure can be modeled as
\begin{equation}
    \mathbf{y} = \mathcal{M} \odot (\mathcal{F}\mathbf{x}) + \eta,
    \label{eq:acquisition_model}
\end{equation}
where $\mathcal{M}$ is a binary sampling mask, $\mathcal{F}$ is the 2D Fourier transform, $\odot$ denotes element-wise multiplication, and $\eta$ is zero-mean complex Gaussian noise. 
Accelerated MRI aims to reconstruct the image $\mathbf{x}$ from the under-sampled measurements $\mathbf{y}$, which is an ill-posed task that requires incorporating prior knowledge~\cite{abu2022image,bora2017compressed,hussein2020image}.

A key challenge in clinical CS is designing undersampling patterns that maximize reconstruction fidelity while minimizing scan time~\cite{safari-review,shimron-review}.
Deep learning approaches tackle this by jointly optimizing a sampling mask and a reconstruction network~\cite{loupe,puert}, replacing hand-crafted sampling rules with data-driven patterns tailored to a specific anatomy. For example, LOUPE~\cite{loupe} learns optimized Cartesian masks in an end-to-end framework, while PUERT~\cite{puert} employs stochastic sampling to improve reconstruction reliability. However, these methods produce a fixed or probabilistic mask for an entire dataset, limiting their ability to adapt to patient-specific anatomical variations in an individual scan~\cite{suno,adasel}.

Scan-adaptive and active methods have been developed to address this limitation~\cite{safari-review,shimron-review}. Methods such as SUNO~\cite{suno} and Ravula et al.~\cite{Ravula2023OptimizingSP} adapt the mask to a given volume from initial measurements but remain static once the trajectory is fixed. In contrast, active sampling updates the acquisition online by selecting future $k$-space measurements conditioned on what has already been acquired~\cite{adasense,adasel}. AdaSense~\cite{adasense} performs zero-shot diffusion posterior sampling and uses posterior variance as an uncertainty signal, while Ada-Sel~\cite{adasel} uses a super-resolution model as a Bayesian uncertainty estimator to assign a mask-reconstruction pair from a finite set of specialist networks. Despite these advances, coupling policy selection with reconstruction can introduce stability issues and reconstruction trade-offs, especially at high acceleration~\cite{adasel}.

Another challenge in MRI reconstruction is evaluation. Prior works primarily report pixel-wise metrics such as PSNR and SSIM; however, these can correlate poorly with radiologist-perceived quality and are sensitive to acquisition noise, especially for methods that emphasize structural fidelity and perceptual realism over strict pixel accuracy~\cite{SSFD}. We therefore complement conventional metrics with Deep Feature Distances (DFDs), including LPIPS~\cite{lpips}, DISTS~\cite{dists}, and Self Supervised-Feature-Distance (SSFD), which have been shown to better capture fine anatomical detail and to exhibit stronger agreement with expert assessment~\cite{SSFD}.

In this work, we use the MedITok tokenizer~\cite{meditok} to define a structured latent space and train a Transformer~\cite{attention-is-all-you-need} to reconstruct image tokens, whose predictive statistics are then used for active sampling policies. \textbf{Our contributions are}: (1) \textbf{Latent Entropy Selection (LES)}, which projects patch-wise token entropy into $k$-space to identify informative sampling lines; (2) \textbf{Gradient-based Entropy Optimization (GEO)}, which selects measurements using the $k$-space gradient of a total latent-entropy objective; and (3) a unified comparative evaluation with retrained baselines on the NYU fastMRI~\cite{fastmri} Knee and Brain datasets at $\times8$ and $\times16$ acceleration, showing improved performance in perceptual and feature-based metrics.

\section{Methodology}

\begin{figure}[!t]
    \includegraphics[width=0.95\textwidth,trim=0 34pt 9pt 0, clip]{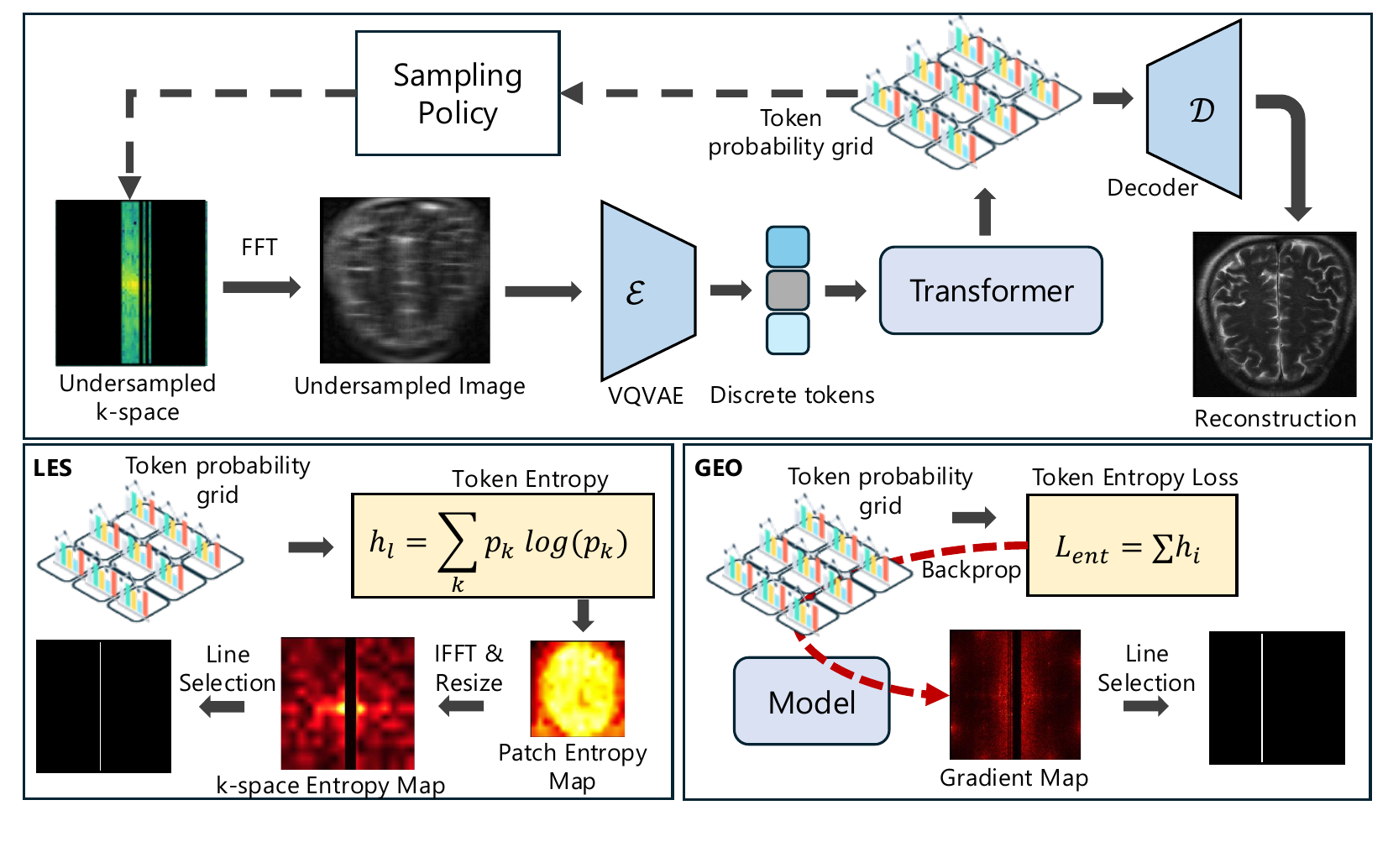}
    \caption{\textbf{Top}: The general pipeline for discrete token prediction and reconstruction. \textbf{Bottom}: The two proposed entropy-driven active sampling policies: Latent Entropy Selection - LES, and Gradient-based Entropy Optimization - GEO.} \label{fig:diagram}
\end{figure}

We formulate active sampling as a sequential decision process in which $k$-space measurements are acquired over multiple time steps. Let $f_\theta(\cdot)$ denote a reconstruction network with parameters $\theta$, trained offline using randomly sampled $k$-space masks to map undersampled measurements to image reconstructions. Given the currently acquired measurements $\mathbf{y}_{t-1}$, an initial reconstruction is obtained as $\mathbf{x}_{t-1} = f_\theta(\mathbf{y}_{t-1})$. At each time step $t$, a policy $\pi$ selects an additional set of sampling locations $\Delta \mathcal{M}_t$ based on this reconstruction. The sampling mask is updated cumulatively according to $\mathcal{M}_t = \mathcal{M}_{t-1} \cup \Delta \mathcal{M}_t$, with $\mathcal{M}_0$ denoting the initial mask.

The policy $\pi$ is optimized to minimize some cost function $\Psi: \mathbb{C}^N\rightarrow \mathbb{R}^+$, under a fixed sampling budget:
\begin{equation}
    \min_{\pi} \; \mathbb{E}_{\mathbf{x} \sim p(\mathbf{x})}
    \left[ 
    % \| \mathbf{x} - f_\theta(\mathbf{y}_{\mathcal{M}_T}) \|_2^2 
    \Psi(f_\theta(\mathbf{y}_{\mathcal{M}_T}), \mathbf{x})
    \right]
    \quad \text{s.t.} \quad \|\mathcal{M}_T\|_0 \leq B,
\end{equation}
where $T$ denotes the final acquisition step and $B$ is a predefined sampling budget. This formulation enables the sampling policy to adapt the $k$-space trajectory to patient-specific anatomical variations during a scan.

\subsection{Reconstruction and Active Sampling}
Given undersampled $k$-space measurements $\mathbf{y}_{\mathcal{M}} \in \mathbb{C}^N$, we first obtain a zero-filled image $\mathbf{x}_{zf} = \mathcal{F}^{-1}(\mathbf{y}_{\mathcal{M}}) \in \mathbb{C}^{H \times W}$, where $N = H \times W$ and $(H,W)$ are the image dimensions, and decompose $\mathbf{x}_{zf}$ into real and imaginary components, $\mathbf{x}_{re}, \mathbf{x}_{im} \in \mathbb{R}^{H \times W}$. 

We utilize the MedITok~\cite{meditok} tokenizer, where an encoder $\mathcal{E}$ produces a latent grid and a quantization operator $Q(.)$ maps each cell to its nearest neighbor in a discrete codebook $\mathcal{Z} = \{\mathbf{z}_k\}_{k=1}^K \subset \mathbb{R}^D$. For a patch size $p$, the image is represented as a sequence of length $L = (H/p) \times (W/p)$, resulting in a quantized embedding $\mathbf{q}_{re} = Q(\mathcal{E}(\mathbf{x}_{re}))$ and  $\mathbf{q}_{im} = Q(\mathcal{E}(\mathbf{x}_{im}))$, where  $\mathbf{q}_{re}, \mathbf{q}_{im} \in \mathbb{R}^{L \times D}$. These streams are integrated using summation and layer normalization to form the initial latent representation $\mathbf{H}_0 = \text{LayerNorm}(\mathbf{q}_{re} + \mathbf{q}_{im})$.
A Transformer decoder $\mathcal{T}{\phi}$ is trained to predict the fully sampled token sequences for the real and imaginary streams, denoted by $\hat{\mathbf{q}}_{re}$ and $\hat{\mathbf{q}}_{im}$, respectively.
% A Transformer decoder $\mathcal{T}_{\phi}$ is trained to predict the fully sampled image tokens. $\mathbf{\hat{q}}_{re}$ and $\mathbf{\hat{q}}_{im}$ corresponding to the fully sampled anatomical features.
The final complex-valued image $\mathbf{\hat{x}} \in \mathbb{C}^{H \times W}$ is recovered via the decoder $\mathcal{D}$ (Fig.~\ref{fig:diagram}, Top):
\begin{equation}
    \mathbf{\hat{x}} = \mathcal{D}(\mathbf{\hat{q}}_{re}) + i \mathcal{D}(\mathbf{\hat{q}}_{im})
\end{equation}

The Transformer predicts a categorical distribution over codebook entries at each latent position. We use the resulting token probabilities to form a spatial uncertainty map, transform it to a $k$-space score map, and select the next line accordingly.

\textbf{Latent Entropy Selection (LES)} uses the predicted token probabilities to guide $k$-space line selection. The Transformer outputs a distribution over the codebook $Z$ for each of the $L$ latent positions. We quantify patch uncertainty via Shannon entropy: $h_l = -\sum_{k=1}^{K} p(z_k | \mathbf{H}_0) \log p(z_k | \mathbf{H}_0)$, where $p(z_k | \mathbf{H}_0)$ is the predicted probability of the $k$-th element. This produces a low-resolution entropy map $\mathbf{h} \in \mathbb{R}^{H/p \times W/p}$, which is bilinearly interpolated to image size $\mathbf{U}_{space} \in \mathbb{R}^{H \times W}$ and transformed to $k$-space via $\mathbf{U}_{kspace} = |\mathcal{F}(\mathbf{U}_{space})|\in  \mathbb{R}^{H \times W}$. Since $k$-space lines correspond to spatial frequency bands, large values in $\mathbf{U}_{kspace}$ indicate frequency content for which the model is most uncertain, making those lines informative to acquire. The next line to be sampled $j^*$ is selected by maximizing the average line amplitude (Fig.~\ref{fig:diagram}, LES):
\begin{equation}
    j^* = \arg\max_{j} \frac{1}{W}\sum_{i=1}^W \mathbf{U}_{kspace}^{(j, i)}
\end{equation}
% \begin{equation} j^* = \arg\max_{j} \text{mean}(\mathbf{U}_{kspace}[j, :]) \end{equation}
% This strategy prioritizes $k$-space lines corresponding to spatial regions where the model is most uncertain.

\textbf{Gradient-based Entropy Optimization (GEO)} identifies informative $k$-space regions by calculating the sensitivity of the total predicted tokens latent entropy $\mathcal{L}_{ent} = \sum_{i=1}^{L} h_i$ with respect to input measurements. As the quantization step in $\mathcal{E}$ is non-differentiable, we employ a Straight-Through Estimator (STE) to backpropagate gradients from the Transformer output through the discrete latent space to the input $k$-space (Fig.~\ref{fig:diagram} GEO). The gradient magnitude map $\mathbf{G} \in \mathbb{R}^{H \times W}$ is computed as $\mathbf{G} = | \partial \mathcal{L}_{ent} / \partial \mathbf{y}_{\mathcal{M}_t} |$, and the next line $j^*$ is selected via:
\begin{equation}
    j^* = \arg\max_{j} \sum_{i=1}^W \mathbf{G}^{(j, i)}
\end{equation}

\section{Experiments}
\subsection{Setup}
\textbf{Datasets.} We evaluate on the NYU fastMRI dataset~\cite{fastmri}. For \textit{knee imaging}, we use the single-coil set ($34K$ training, $7K$ testing slices) center-cropped to $320 \times 320$~\cite{fastmri}. For \textit{brain imaging}, we select a multi-coil subset (57762 training, 240 testing slices) and emulate single-coil (ESC) data following~\cite{Tygert2018SimulatingSM}, cropped to $256 \times 256$~\cite{fastmri}. The testing set was chosen according to~\cite{suno}. We use 1D Cartesian masks with a 4\% center fraction $\rho_c = 0.04$~\cite{fastmri,SSFD}. The non-central sampling budget is $B = \text{round}(N(1 - \rho_c) / R)$, where $N$ and $R$ denote resolution and acceleration.

\textbf{Metrics.} We assess quality via pixel-wise metrics (PSNR, SSIM, NMSE) and Deep Feature Distances (LPIPS, DISTS, SSFD)~\cite{SSFD}. Metrics are computed per volume/scan and averaged over the test set~\cite{SSFD}. All metrics were computed using the publicly available implementation provided by~\cite{SSFD}.

\textbf{Implementation Details.} The decoder-only Transformer $\mathcal{T}_{\phi}$ has $N{=}24$ layers, $H{=}16$ self-attention heads, embedding dimension $d{=}1024$, and patch size $p{=}16$. Training minimizes a token-level cross-entropy loss over codebook indices. The MedITok tokenizer~\cite{meditok} is frozen and uses a codebook of size $|\mathcal{Z}|{=}32768$. Training was performed on an NVIDIA RTX A6000 for 100 epochs using the AdamW optimizer with a learning rate of $1 \times 10^{-4}$ and batch size of 32~\cite{SSFD}. Policies $\pi$ select vertical phase-encoding lines on 1D Cartesian masks.

\begin{table}[!t]
\centering
\caption{Comparison on the fastMRI Knee and Brain (ESC) datasets. R denotes acceleration factor.}
\label{tab:consolidated_results}
\footnotesize % Increased font size for better readability
\resizebox{\columnwidth}{!}{%
\begin{tabular}{@{}llll cccccc@{}}
\toprule
\textbf{Dataset} & \textbf{R} & \textbf{Method} & \textbf{Model} & \textbf{PSNR} $\uparrow$ & \textbf{SSIM} $\uparrow$ & \textbf{NMSE} $\downarrow$ & \textbf{LPIPS} $\downarrow$ & \textbf{DISTS} $\downarrow$ & \textbf{SSFD} $\downarrow$ \\ \midrule
%%%%%%%%% X8 KNEE %%%%%%%%
\multirow{15}{*}{\rotatebox[origin=c]{90}{Knee}} 
& \multirow{7}{*}{$\times 8$} 
& Random & U-Net & 31.28 & 0.7250 & 0.0368 & 4.47 & 0.24 & 14.35 \\
&& LOUPE~\cite{loupe} & U-Net & \underline{32.21} & \underline{0.7459} & \underline{0.0317} & 5.26 & 0.31 & 9.41 \\
&& PUERT~\cite{puert} & ISTA-Unfold~\cite{ista} & \textbf{33.63} & \textbf{0.7963} & \textbf{0.0232} & 5.25 & 0.32 & \underline{9.16} \\
&& Ada-sel~\cite{adasel} & VarNet$\times3$~\cite{varnet} & 31.23 & 0.7396 & 0.0385 & 4.29 & \underline{0.21} & 11.58 \\
&& AdaSense~\cite{adasense} & DDRM~\cite{ddrm} & 31.19 & 0.6552 & 0.0420 & 4.68 & 0.24 & 11.63 \\
&& \textbf{LES (Ours)} & \textbf{Transformer} & 30.29 & 0.6490 & 0.0478 & \underline{3.70} & \textbf{0.15} & \textbf{7.35} \\
&& \textbf{GEO (Ours)} & \textbf{Transformer} & 30.26 & 0.6498 & 0.0478 & \textbf{3.66} & \textbf{0.15} & \textbf{7.35} \\ \cmidrule{2-10}
& \multirow{7}{*}{$\times 16$}
%%%%%%%%% X16 KNEE %%%%%%%%
& Random & U-Net & 30.05 & 0.6863 & 0.0463 & 5.49 & 0.30 & 17.03 \\
&& LOUPE~\cite{loupe} & U-Net & \textbf{31.09} & \textbf{0.7199} & \textbf{0.0370} & 6.08 & 0.35 & \underline{13.91} \\
&& PUERT~\cite{puert} & ISTA-Unfold~\cite{ista} & \underline{31.08} & \underline{0.7118} & \underline{0.0372} & 6.43 & 0.37 & 14.75 \\
&& Ada-sel~\cite{adasel} & VarNet$\times3$~\cite{varnet} & 29.49 & 0.6901 & 0.0530 & 5.42 & 0.28 & 15.02 \\
&& AdaSense~\cite{adasense} & DDRM~\cite{ddrm} & 30.05 & 0.6061 & 0.0512 & 4.78 & \underline{0.24} & 14.29 \\
&& \textbf{LES (Ours)} & \textbf{Transformer} & 28.59 & 0.5980 & 0.0655 & \underline{4.12} & \textbf{0.18} & \textbf{8.82} \\
&& \textbf{GEO (Ours)} & \textbf{Transformer} & 28.61 & 0.5983 & 0.0653 & \textbf{4.11} & \textbf{0.18} & \textbf{8.82} \\ \cmidrule{2-10}
& Oracle & \textit{VQVAE} & \textit{VQVAE} & 32.14 & 0.7111 & 0.0362 & 2.65 & 0.07 & 4.57 \\ \midrule
%%%%%%%%% X8 BRAIN %%%%%%%%
\multirow{15}{*}{\rotatebox[origin=c]{90}{Brain}} 
& \multirow{7}{*}{$\times 8$} 
& Random & U-Net & 27.83 & 0.7313 & 0.0361 & 3.78 & 0.23 & 10.91 \\
&& LOUPE~\cite{loupe} & U-Net & 28.80 & \underline{0.7869} & 0.0287 & 4.26 & 0.27 & 7.87 \\
&& PUERT~\cite{puert} & ISTA-Unfold~\cite{ista} & \textbf{30.73} & \textbf{0.8219} & \textbf{0.0185} & 4.33 & 0.29 & 7.15 \\
&& Ada-sel~\cite{adasel} & VarNet$\times3$~\cite{varnet} & \underline{29.44} & 0.7731 & \underline{0.0253} & 3.65 & 0.22 & 8.95 \\
% && AdaSense~\cite{adasense} & DDRM~\cite{ddrm} & - & - & - & - & - & - \\
%&& Random (Tr.) & \textbf{Transformer} & 25.99 & 0.6387 & 0.0549 & 3.26 & 0.14 & 7.49 \\
&& \textbf{LES (Ours)} & \textbf{Transformer} & 27.11 & 0.6702 & 0.0424 & \underline{3.07} & \underline{0.14} & \underline{6.76} \\
&& \textbf{GEO (Ours)} & \textbf{Transformer} & 27.26 & 0.6783 & 0.0407 & \textbf{2.96} & \textbf{0.13} & \textbf{6.67} \\ \cmidrule{2-10}
& \multirow{7}{*}{$\times 16$}
%%%%%%%%% X16 BRAIN %%%%%%%%
& Random & U-Net & 25.96 & 0.6748 & 0.0556 & 4.61 & 0.27 & 13.44 \\
&& LOUPE~\cite{loupe} & U-Net & \textbf{28.13} & \textbf{0.7460} & \textbf{0.0332} & 4.93 & 0.30 & 10.39 \\
&& PUERT~\cite{puert} & ISTA-Unfold~\cite{ista} & \underline{27.69} & \underline{0.7291} & \underline{0.0369} & 5.18 & 0.31 & 11.53 \\
&& Ada-sel~\cite{adasel} & VarNet$\times3$~\cite{varnet} & 26.80 & 0.7144 & 0.0457 & 4.51 & \underline{0.26} & 11.55 \\
% && AdaSense~\cite{adasense} & DDRM~\cite{ddrm} & - & - & - & - & - & - \\
%&& Random (Tr.) & \textbf{Transformer} & 24.50 & 0.5901 & 0.0779 & 3.48 & \textbf{0.16} & 8.70 \\
&& \textbf{LES (Ours)} & \textbf{Transformer} & 24.59 & 0.5904 & 0.0759 & \underline{3.50} & \textbf{0.16} & \underline{8.37} \\
&& \textbf{GEO (Ours)} & \textbf{Transformer} & 24.75 & 0.5952 & 0.0731 & \textbf{3.48} & \textbf{0.16} & \textbf{8.26} \\ \cmidrule{2-10}
& Oracle & \textit{VQVAE} & \textit{VQVAE} & 29.48 & 0.7310 & 0.0251 & 2.22 & 0.08 & 4.28 \\
\bottomrule
\end{tabular}}
\end{table}

\subsection{Results}
We evaluate our proposed LES and GEO policies against baselines: LOUPE \cite{loupe}, PUERT \cite{puert}, AdaSense \cite{adasense} and Ada-Sel~\cite{adasel}. All methods were evaluated on the fastMRI Knee and Brain datasets at acceleration factors of $\times8$ and $\times16$. Unless noted otherwise, baselines were retrained and evaluated under a unified protocol on the same use-cases for fair comparison. AdaSense brain results are omitted because no brain experiments or weights are provided; we did not retrain a diffusion baseline.

\textbf{Quantitative Performance.} Table \ref{tab:consolidated_results} shows that baselines such as PUERT consistently achieve higher PSNR and SSIM than our methods. For example, on fastMRI Knee at $\times8$, PUERT reaches 33.63 dB PSNR (vs.\ $\sim$30.2 dB for LES/GEO), and AdaSense also maintains higher PSNR (31.19 dB). A similar trend is seen across Knee/Brain at both $\times8$ and $\times16$ accelerations, while our policies consistently achieve the best perceptual and feature-based metrics (LPIPS, DISTS, SSFD). On fastMRI Knee at $\times16$, LES and GEO achieve an SSFD of 8.82, outperforming PUERT (14.75; \textbf{40.2\% reduction}), LOUPE (13.91; \textbf{36.6\% reduction}), and AdaSense (14.29; \textbf{38.3\% reduction}). At Knee $\times8$, LES/GEO achieve LPIPS scores of $\sim$3.70, improving over PUERT (5.25; \textbf{$\sim$29.5\% lower}) and AdaSense (4.68; \textbf{$\sim$20.9\% lower})~\cite{puert,adasense}. In the Brain dataset, GEO reaches the best SSFD of 6.67 at $\times8$, improving over PUERT (7.15; \textbf{6.7\% lower}) and Ada-Sel (8.95; \textbf{25.5\% reduction}).

\begin{figure}[!t]
    \centering
    \includegraphics[width=0.85\textwidth]{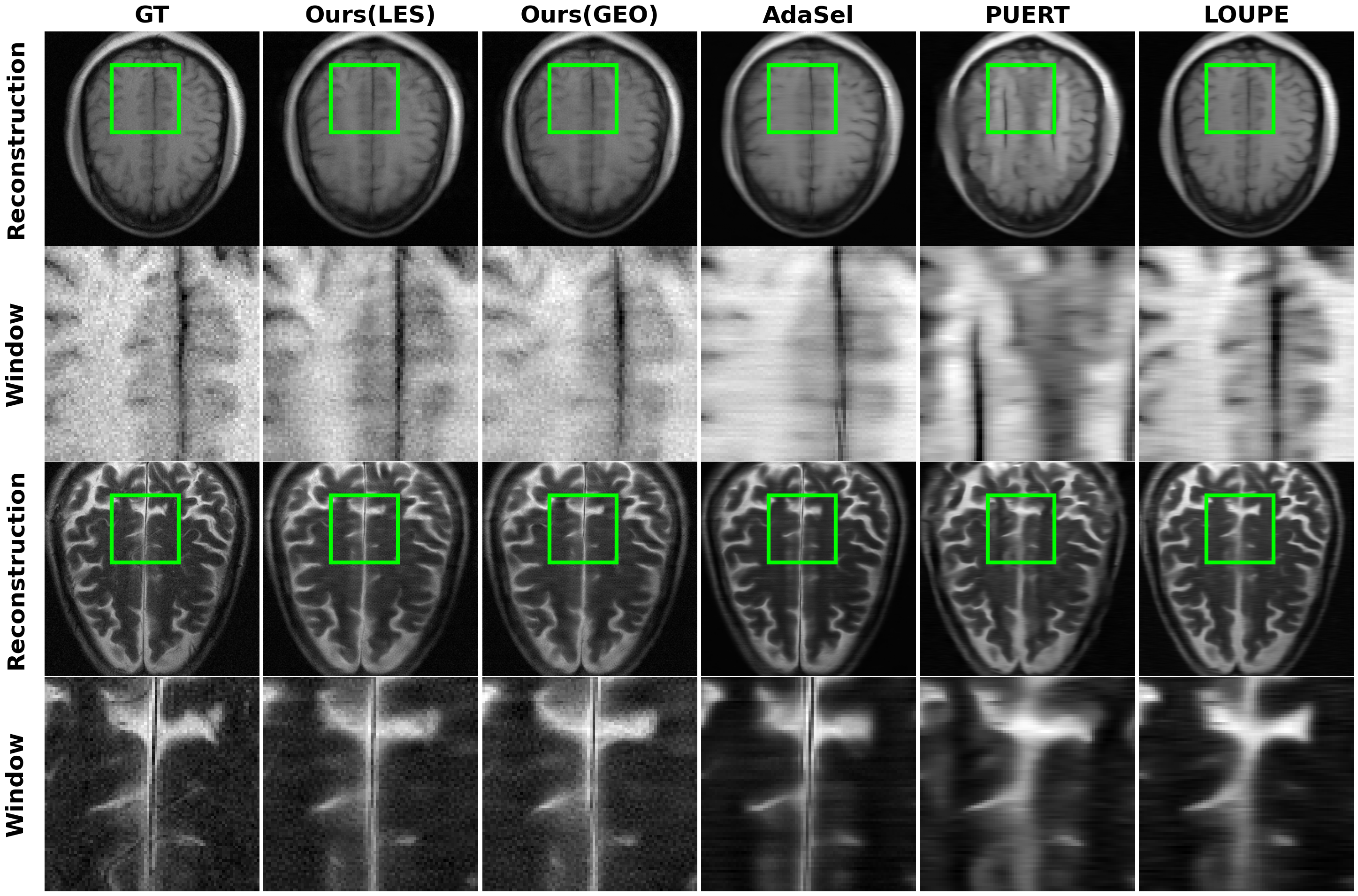}\\
    \scriptsize (a) fastMRI Brain dataset~\cite{fastmri}.\\ % [0.5ex] is a minor vertical adjustment, not a prohibited \vspace
    \includegraphics[width=0.85\textwidth]{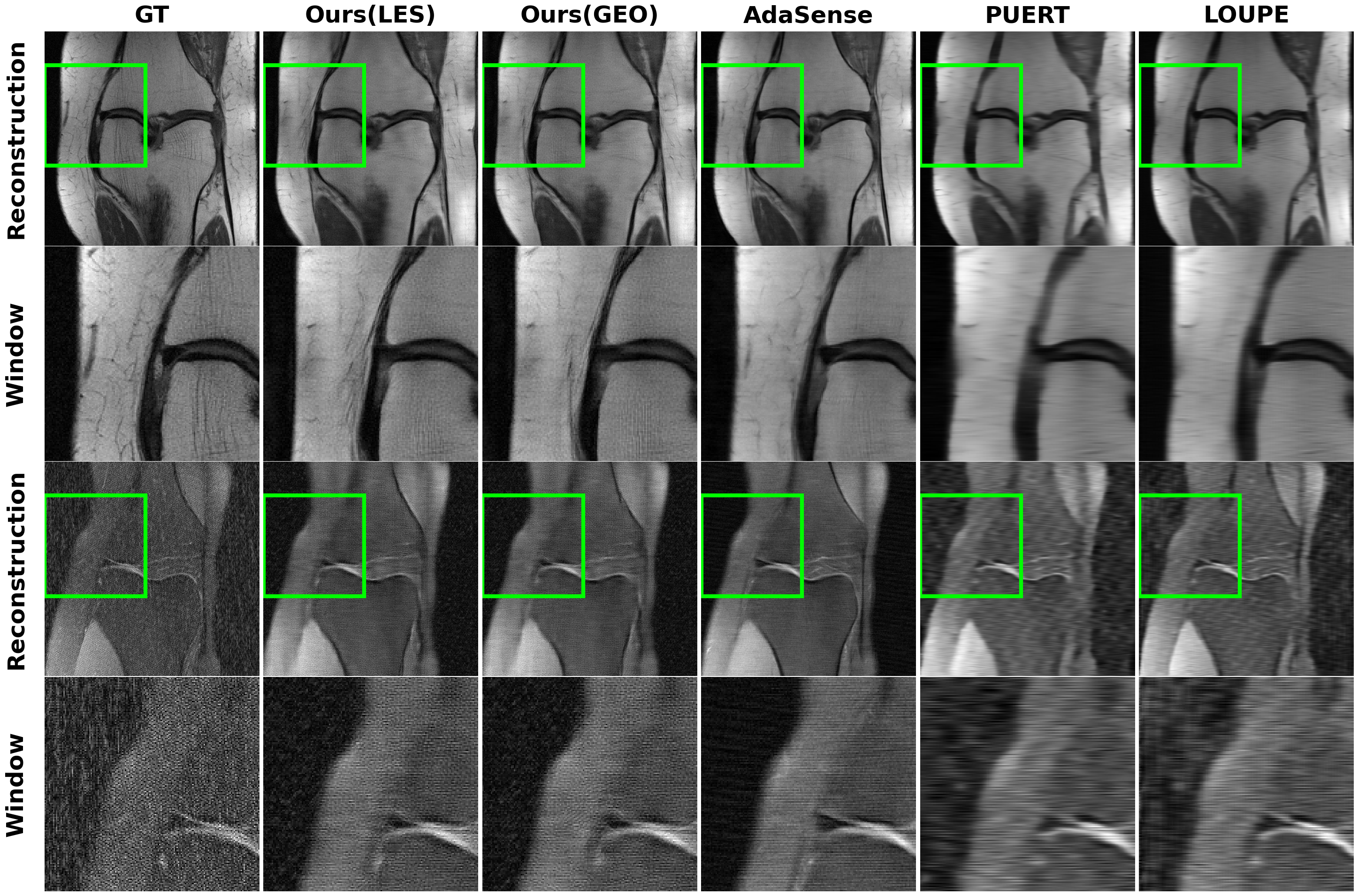}\\
    \scriptsize (b) fastMRI Knee dataset~\cite{fastmri}.
    \caption{Qualitative comparison of reconstruction results at $\times 16$ acceleration.}
    \label{fig:merged_results2}
\end{figure}

\textbf{Qualitative Analysis.} Visual comparisons in Fig.~\ref{fig:merged_results2} are consistent with the quantitative trends. At $\times16$, LES and GEO preserve details and local texture better than LOUPE~\cite{loupe} and PUERT~\cite{puert}, which show stronger over-smoothing. As shown in Fig.~\ref{fig:merged_results2}(b), some ground-truth images contain acquisition noise that our method tends to suppress. This produces visually cleaner reconstructions, but can increase pixel-wise deviation relative to ground truth.

\textbf{Oracle.} We include a VQ-VAE Oracle based on the frozen MedITok tokenizer~\cite{meditok} to estimate the upper bound of the discrete latent space, by directly encoding--decoding the ground-truth image. The Oracle achieves strong distortion-based performance (e.g., Knee $\times8$: 32.14 dB PSNR, 0.7111 SSIM) and the best DFD scores (LPIPS 2.65, DISTS 0.07, SSFD 4.57), while LES/GEO remain closer to the Oracle in DFD metrics than in PSNR/SSIM, suggesting that their main gap is in pixel-level fidelity rather than perceptual/anatomical structure.

\textbf{Runtime and Efficiency.} Table~\ref{tab:efficiency} summarizes latency and throughput at $\times16$ acceleration. LES provides the best throughput (0.97 fps), while GEO is slower but remains far faster than prior active baselines.

\textbf{Ablation Study}. To isolate the effects of the Latent Transformer and iterative active sampling, we ablate the fastMRI Brain (ESC) setting at $\times8$ acceleration (Table~\ref{tab:ablation}). Both LES and GEO improve over random sampling even with a single acquisition step ($T=1$), with the largest gains in feature-based metrics (e.g., LES improves SSFD from 7.49 to 6.85 and DISTS from 0.1421 to 0.1380). Increasing the number of active steps to $T=22$ further improves performance, with GEO achieving the best overall results (DISTS 0.1315, SSFD 6.6656).

\begin{table}[!t]
\centering
\caption{Ablation study at $\times 8$ acceleration on the fastMRI Brain (ESC) set. $T$ denotes sampling steps.}
\label{tab:ablation}
\fontsize{8}{9}\selectfont
\begin{tabular}{@{}lccccccc@{}}
\toprule
\textbf{Sampling Policy} & \textbf{$T$} & \textbf{PSNR $\uparrow$} & \textbf{SSIM $\uparrow$} & \textbf{LPIPS $\downarrow$} & \textbf{DISTS $\downarrow$} & \textbf{SSFD $\downarrow$} \\ \midrule
Random & 0 & 26.00 & 0.6387 & 3.2587 & 0.1421 & 7.4903 \\
LES & 1 & 27.07 & 0.6697 & 3.0733 & 0.1380 & 6.8490 \\
GEO & 1 & 26.95 & 0.6676 & \underline{3.0656} & 0.1381 & 6.9482 \\
LES & 22 & \underline{27.11} & \underline{0.6702} & 3.0730 & \underline{0.1379} & \underline{6.7607} \\ 
GEO & 22 & \textbf{27.26} & \textbf{0.6783} & \textbf{2.9630} & \textbf{0.1315} & \textbf{6.6656} \\ \bottomrule
\end{tabular}
\end{table}

\begin{table}[!t]
\centering
\caption{Computational efficiency analysis at $\times16$ acceleration. Total time includes policy execution and final reconstruction. Latency is measured per active sampling step. $T$ denotes sampling steps.}
\label{tab:efficiency}
\fontsize{8}{9}\selectfont
\begin{tabular}{@{}l c c c c@{}}
\toprule
\textbf{Method} & \textbf{$T$} & \textbf{Step Latency} & \textbf{Total Time} & \textbf{Throughput (fps)} \\ \midrule
AdaSense~\cite{adasense} & 8 & 8.49 $\pm$ 0.01 s & 76.63 $\pm$ 0.12 s & 0.01 \\
Ada-Sel~\cite{adasel}   & 1 & 5.43 $\pm$ 1.32 s &  5.45 $\pm$ 0.14 s & 0.18 \\ \midrule
Random (ours)        & 0 & --- & 112.8 $\pm$ 2.0 ms & 8.86 \\
\textbf{LES (Ours)}    & 8 & 229.1 $\pm$ 17.0 ms & 1.04 $\pm$ 0.08 s & 0.97 \\
\textbf{GEO (Ours)}    & 8 & 350.3 $\pm$ 6.1 ms & 2.02 $\pm$ 0.03 s & 0.50 \\
\bottomrule
\end{tabular}
\end{table}

\section{Discussion}
\textbf{Perception-Distortion Trade-off.} Our results suggest that discrete latent uncertainty changes the sampling/reconstruction behavior from optimizing pixel-wise fidelity to preserving anatomically plausible structure. Distortion-oriented baselines such as PUERT~\cite{puert} and LOUPE~\cite{loupe} achieve higher PSNR/SSIM, but produce over-smoothed reconstructions that can suppress fine anatomical details and diagnostically relevant texture. The PSNR gap, which reaches up to ~3 dB in certain settings, is a recognized limitation of our approach. A contributing factor is visible acquisition noise in some ground-truth images. By not reproducing this noise, our model incurs higher pixel-wise error. However, this is not the only contributing factor. Comparisons with the VQ-VAE Oracle indicate that the discrete latent representation itself is not the primary performance bottleneck, but the Transformer’s current predictive ability to fully recover the latent sequence.
\textbf{Clinical Utility.} Quantitative and qualitative results show that LES and GEO often preserve boundaries and local structure more effectively. In the qualitative examples, our reconstructions are typically sharper and partially suppress acquisition noise. From a deployment perspective, LES is substantially more efficient than prior active methods, achieving 0.97 fps.
Overall, LES and GEO achieve comparable quality, with GEO being slightly better, while LES offers nearly $2\times$ higher throughput.

\textbf{Future Work} will focus on will focus on improving latent-token prediction and study hybrid losses, and extend to multi-coil data and non-Cartesian trajectories with alternative complex tokenizations (e.g., magnitude/phase).

 %% removed for anonymized MICCAI submission.
    
    % The following acknowledgement and disclaimer sections can be removed for the double-blind review process.  If and when your paper is accepted, reinsert the acknowledgement and the disclaimer clause in your final camera-ready version.
    % IF you opted to include the acknowledgement and disclaimer sections, they will count towards the 8-page limit.

\begin{credits}
\subsubsection{\ackname} This work was partially supported by a grant from The Center for AI and Data Science at Tel Aviv University (TAD).

% \subsubsection{\discintname}
% The authors declare no competing interests
\end{credits}

%
% ---- Bibliography ----
%
% BibTeX users should specify bibliography style 'splncs04'.
% References will then be sorted and formatted in the correct style.
%
% \newpage
\FloatBarrier
\bibliographystyle{splncs04}
\bibliography{refs}
%
% \begin{thebibliography}{8}
% \bibitem{ref_article1}
% Author, F.: Article title. Journal \textbf{2}(5), 99--110 (2016)

% \bibitem{ref_lncs1}
% Author, F., Author, S.: Title of a proceedings paper. In: Editor,
% F., Editor, S. (eds.) CONFERENCE 2016, LNCS, vol. 9999, pp. 1--13.
% Springer, Heidelberg (2016). \doi{10.10007/1234567890}

% \bibitem{ref_book1}
% Author, F., Author, S., Author, T.: Book title. 2nd edn. Publisher,
% Location (1999)

% \bibitem{ref_proc1}
% Author, A.-B.: Contribution title. In: 9th International Proceedings
% on Proceedings, pp. 1--2. Publisher, Location (2010)

% \bibitem{ref_url1}
% LNCS Homepage, \url{http://www.springer.com/lncs}, last accessed 2023/10/25
% \end{thebibliography}
\end{document}